\documentclass{article}

\usepackage[final]{colm2026_conference}

\usepackage{microtype}
\usepackage{graphicx}
\usepackage{booktabs} 
\usepackage[most]{tcolorbox}
\usepackage{hyperref}

\usepackage{lineno}
\definecolor{darkblue}{rgb}{0, 0, 0.5}
\hypersetup{colorlinks=true, citecolor=darkblue, linkcolor=darkblue, urlcolor=darkblue}

\usepackage{amsmath}
\usepackage{amssymb}
\usepackage{mathtools}
\usepackage{amsthm}

\usepackage{algorithm}
\usepackage{algorithmic}

\usepackage[capitalize,noabbrev]{cleveref}

\theoremstyle{plain}

\theoremstyle{definition}

\theoremstyle{remark}

\usepackage[disable,textsize=tiny]{todonotes}

\usepackage{graphicx}
\usepackage{subcaption}

\usepackage{caption}
\begin{document}

\title{Idea Search: Guiding Tree Search with Ideas to Explore \\ Diverse Scientific Methods}

\author{Xuefei (Julie) Wang\thanks{Work done during an internship at Google Research.} \\
California Institute of Technology \\
\texttt{xwang3@caltech.edu} \\
\And
Hao Cui \\
Google Research \\
\And
Michael P. Brenner \\
Google Research \& Harvard University \\
\And
Subhashini Venugopalan \\
Google Research \\
}

\ifcolmsubmission
\linenumbers
\fi

\maketitle

\begin{abstract}

Tree Search-based test-time scaling of LLMs is a powerful tool for automated scientific coding. However, pure Tree Search sometimes struggles with systematic exploration, becoming trapped in local optima, or unproductive loops, especially in the vast search space of scientific methods. To address this limitation, we propose \textbf{Idea Search}, a framework that systematically integrates a dynamic ``Idea Bank'' into Tree Search. Idea Search involves three steps: (1) decomposing existing methods into atomic ideas, (2) sampling from this bank of ideas to guide branches of code mutations, and (3) dynamically updating the bank with new ideas discovered through execution. On single-cell RNA-sequencing (scRNA-seq) batch integration, Idea Search reliably breaks the plateau of a strong pure Tree Search baseline, improving the mean score from 0.678 to 0.697 and reaching a best score of 0.728. We then characterize which design choices drive these gains: bank augmentation helps bandit sampling but not random sampling, ``Exploratory'' prompting that prioritizes new ideas surfaces the rare best-performing solutions, while increasing sampling-level exploration is counterproductive.
\end{abstract}

\section{Introduction}
\label{introduction}

Advanced test-time scaling techniques, such as Tree Search, enable Large Language Models (LLMs) to powerfully automate complex code generation in machine learning and research engineering~\citep{aygun2025ai, jiang2025aide, toledo2025ai}. Beyond simply hill-climbing on a predefined metric, these systems have demonstrated the capacity for genuine scientific method discovery, signaling a shift from simple code optimization to methodological innovation. For instance, one such system discovered single-cell data analysis methods that outperform established, human-developed tools~\citep{aygun2025ai}.

Although Tree Search has proven to be a powerful strategy, successful applications also reveal a core limitation: when left unguided, the search can struggle with systematic exploration—sometimes entering unproductive loops~\citep{antoniades2024swe} or getting trapped in greedy local optima~\citep{jiang2025aide}. To achieve maximum performance, especially in vast idea spaces like science, Tree Search will often require external intervention or specific guidance~\citep{aygun2025ai}. This guidance often takes the form of human-provided research strategies, including reproductions of established expert methods, recombinations of multiple techniques, or ideas sourced from systems such as Deep Research~\citep{deepresearch_g, deepresearch_o}. While effective, these interventions are not automated and do not explore ideas systematically, often requiring many isolated Tree Searches anchored to different initial methods.

This work addresses this gap. An expert method is rarely monolithic: it can be decomposed into small, recombinable design choices, for example single methodological steps like ``condition the decoder on batch ID'' or ``merge per-batch nearest-neighbor sets.'' We refer to such an atomic, executable unit as an \emph{idea}, and treat a method as a particular combination of ideas that a search can mix and recombine. We investigate whether a more automated and systematic method for exploring this space of ideas can be integrated directly into the code optimization process to uncover novel and performant methods in a single, unified Tree Search and to explore various ideas simultaneously. Our approach, which we term \textbf{Idea Search}, builds upon the framework of~\citep{aygun2025ai} and centers on three key steps: (1) decomposing existing, high-quality methods into a bank of ideas; (2) sampling from this ``Idea Bank'' to guide the LLM's mutation proposals during the Tree Search; and (3) dynamically updating the bank with new, high-scoring concepts discovered during the search process.

\begin{figure}[t]
    \centering
    \begin{subfigure}[b]{0.49\columnwidth}
            \centering
            \includegraphics[width=\linewidth]{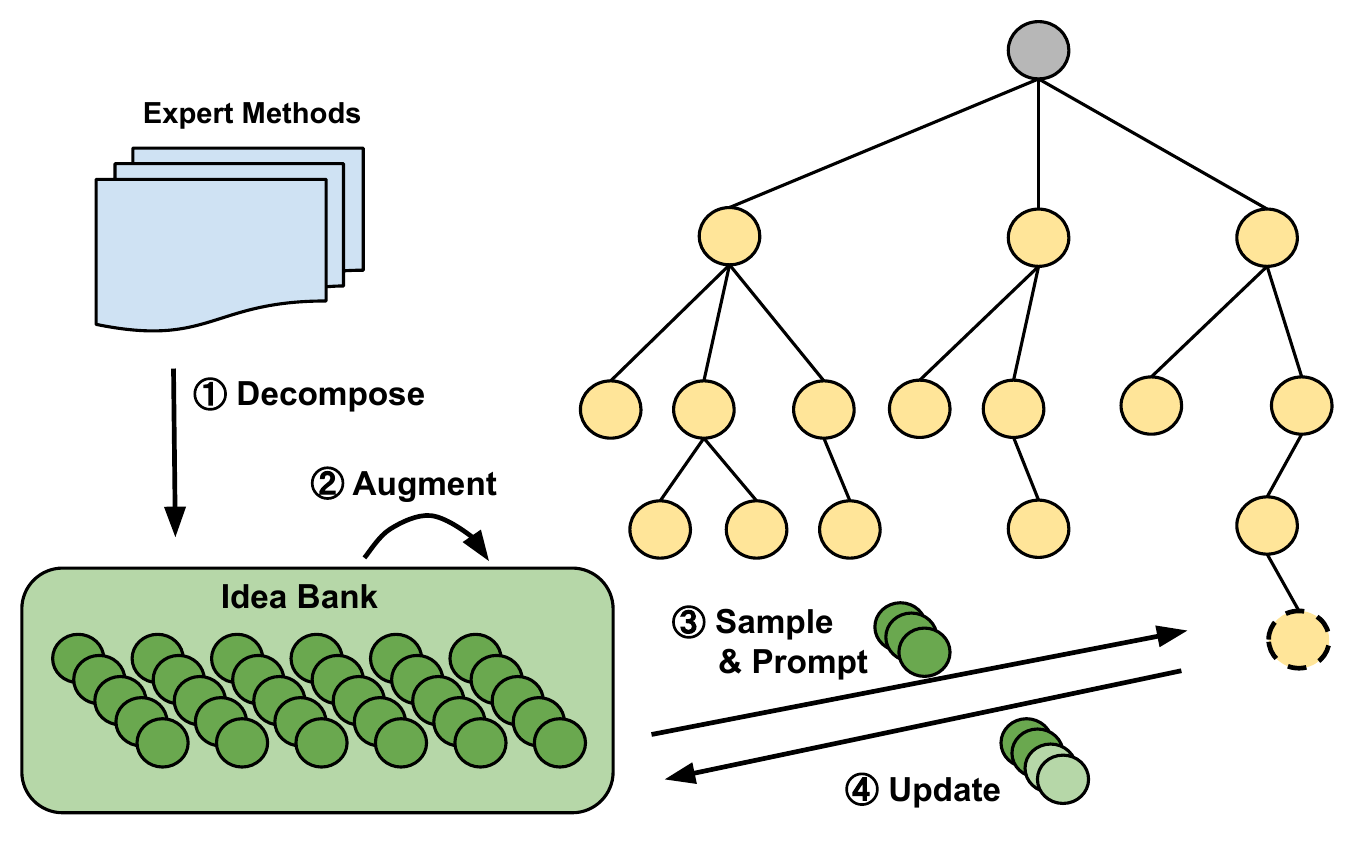}
            \caption{Idea Search Procedure.}
            \label{fig:idea_search_top}
        \end{subfigure}
        \hfill
    \begin{subfigure}[b]{0.49\columnwidth}
            \centering
            \includegraphics[width=\linewidth]{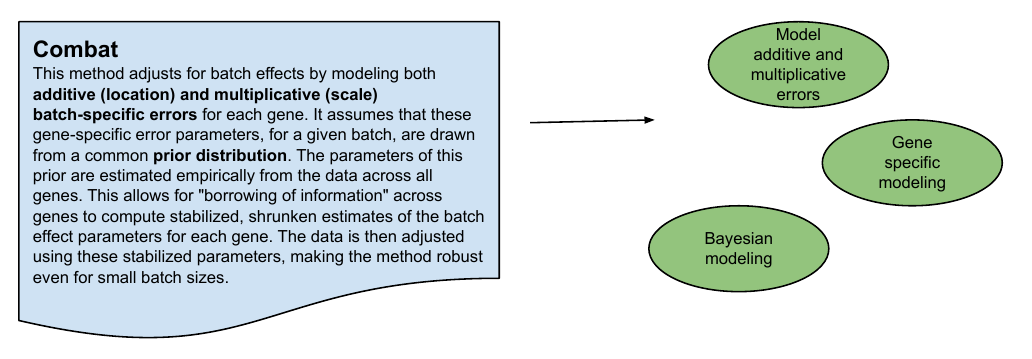}
            \caption{Example of Decomposing an Expert Method into Ideas.}
            \label{fig:idea_search_bottom}
        \end{subfigure}

        \caption{\textbf{Idea Search couples a dynamic Idea Bank with Tree Search.} Expert methods are decomposed into reusable ideas that guide code mutations and are dynamically updated as new solutions are discovered.}
        \label{fig:overview}
        \vspace{.5cm}
\end{figure}

To validate our approach, we apply it to the single-cell RNA-sequencing (scRNA-seq) batch integration problem~\citep{xu2023automatic}. This domain serves as an ideal and challenging testbed for automated scientific discovery, as demonstrated by several works~\citep{aygun2025ai, chung2025station}. Specifically, \citet{aygun2025ai} showed that a pure Tree Search method, even without injecting external ideas, already establishes a highly competitive baseline, conceptually similar to ComBat~\citep{johnson2007adjusting}. With this strong baseline, we investigate whether Idea Search can unlock further performance gains and achieve results beyond what a pure Tree Search alone can find. In summary, our key contributions are as follows:
\begin{itemize}
    \item We propose and formalize Idea Search, a framework that integrates a dynamic Idea Bank directly into the Tree Search loop to systematically guide the LLM's conceptual exploration of candidate methods.
    \item We demonstrate that the Idea Search framework reliably breaks the performance plateau established by a strong pure Tree Search baseline in the challenging domain of scRNA-seq batch integration.
    \item We characterize how the main design choices of Idea Search interact: sampling strategy (Bandit vs.\ Random), bank augmentation (Expert-Only vs.\ Augmented, Fig.~\ref{fig:idea_search_top}), and prompting strategy (Conservative vs.\ Exploratory). We find their effects are interdependent rather than uniform: bank augmentation helps bandit but not random sampling, ``Exploratory'' prompting that prioritizes new ideas surfaces the rare best-performing solutions, and more sampling-level exploration is counterproductive.
\end{itemize}


\section{Related Work}
\label{related_work}


\textbf{Test-Time Scaling, Tree Search, and Evolutionary Systems:} Test-time scaling, which improves model performance by allocating more computation at inference time, is a critical research direction for LLMs. These strategies range from ensembling multiple sampled outputs (e.g., self-consistency~\citep{wang2022self}) to more complex, guided exploration methods~\citep{yao2023tree, yao2022react, novikov2025alphaevolve}. Among the most effective techniques is Tree Search, which has become an increasingly popular method for tackling complex coding and engineering problems~\citep{antoniades2024swe, jiang2025aide, toledo2025ai}, and, most relevant to our work, scientific software development and discovery~\citep{aygun2025ai}. Recently, this paradigm has been extended to adaptive evolutionary and self-evolving agent frameworks that optimize strategies, parameters, or codebases at test time, whether through meta-evolution and generative optimization~\citep{liu2026evox, chi2026frontier}, adaptive zeroth-order search~\citep{cemri2026adaevolve}, or test-time training and reinforcement learning~\citep{wang2025thetaevolve, yuksekgonul2026learning}. These systems improve the generator itself—its weights, parameters, or search policy—rather than maintaining an explicit, human-readable account of the concepts driving each improvement. Idea Search instead couples the code search with a decomposed, dynamically updated bank of such concepts, making conceptual exploration a first-class part of the loop.

However, in practice, Tree-Search-based systems are susceptible to performance plateaus and may get stuck in repetitive, unproductive loops~\citep{antoniades2024swe} or greedy local optima~\citep{jiang2025aide}. While ML-Master~\citep{liu2025ml} addresses this by guiding code generation with reasoning, it is constrained to the local context of a node's siblings and parent. Moreover, the findings of \citet{aygun2025ai} demonstrated that providing the Tree Search with external ideas often improves upon an unguided Tree Search, yet these strategies remain static over time. To address these limitations, we propose a global, dynamic Idea Bank to guide code mutation and systematically explore the idea space.

\textbf{LLM-based Research Ideation:} The use of LLMs for research idea generation has gained significant momentum. Much of this work focuses on maximizing the novelty of ideas, often employing iterative designs that involve literature review, knowledge graph construction, and chain-of-ideas~\citep{baek2024researchagent, wang2024scimon, li2024chain, hu2024nova}. However, while novelty can be reliably scored by other LLMs or human judges, a critical challenge remains: the feasibility and practical utility of these generated ideas are inherently difficult to assess without actual implementation and testing. 

Indeed, recent studies suggest a significant ``ideation-execution gap.'' LLM-generated ideas, while often rated as highly novel, are frequently found to be weaker on feasibility and practical execution~\citep{si2024can, si2025ideation}. This suggests that a more viable path, and the one we explore in this paper, is an execution-verified research ideation system, where ideas are immediately compiled, executed, and scored within a rigorous optimization framework, bridging the gap between a novel concept and a functional, high-performance solution.


\textbf{Searching the Idea Space during Code Generation:} The value of exploring in the natural language ``idea space'', rather than in the code space alone, has been proven effective. For instance, Self-planning~\citep{jiang2024self} allowed a decomposition of complex intent before implementation. PlanSearch~\citep{wang2024planning} explicitly searches this natural language plan space to encourage solution diversity. However, these methods explore the plan space \emph{open-loop}: the plan is generated before any code runs and is never revised by how the resulting code performs. Idea Search instead closes this loop. Each idea carries an execution-verified score and that score steers which ideas are sampled and prompted in subsequent mutations, so the bank's conceptual understanding is continually updated from empirical results rather than fixed in a single planning step.

\textbf{Connection to Reinforcement Learning:} Selecting ideas from the dynamic Idea Bank to guide mutations can be framed as a Markov Decision Process (MDP): the current codebase is the state, selecting an idea is the action, and the change in execution performance is the reward. Our bandit-based UCB sampling is then a simplified, model-free RL algorithm on this MDP---a lightweight, weight-free alternative to the explicit reinforcement learning or weight fine-tuning used in recent test-time and self-evolving frameworks~\citep{wang2025thetaevolve, yuksekgonul2026learning}.

\section{Method}
\label{method}

Our method integrates a systematic search over ideas directly into the Tree-Search optimization loop for code mutation. The core of our framework consists of an Idea Bank that is built, sampled from, and dynamically updated during the optimization process (Fig.~\ref{fig:overview}); Algorithm~\ref{alg:idea-search} states the full loop.

\textbf{Idea Bank:} We first construct an initial Idea Bank by curating a set of existing, high-quality ``expert'' methods for the target domain. Each method (represented as a text summary) is decomposed into its core conceptual components or ideas. These ideas are stored as short textual descriptions in the bank (Fig.~\ref{fig:idea_search_bottom}). Optionally, we ask an LLM (Gemini 2.5 Pro) to augment the Idea Bank by proposing more similar ideas.

\textbf{Idea-Guided Node Mutation:} During the Tree Search optimization process, at each mutation step, we sample one or more ideas from the bank. These ideas are then inserted into the LLM’s code mutation prompt. This prompt instructs the LLM to modify the current solution (i.e., the current node in the search tree) to incorporate the sampled idea.

\textbf{Dynamic Bank and Score Updates:} The framework is dynamic. Whenever a newly generated solution is evaluated, it is fed back into the decomposition-and-update process. An LLM summarizes this new solution into its constituent ideas, which are then added to the bank, progressively expanding the conceptual search space. Simultaneously, the solution's score is used to update the statistics of all ideas it contains; an idea's score is maintained as the average score of all solutions generated using that specific idea.

\begin{algorithm}[t]
\caption{Idea Search}
\label{alg:idea-search}
\begin{algorithmic}[1]
\REQUIRE Expert method summaries $\mathcal{M}$; sampling strategy (Random or Bandit); mutation budget $b$
\ENSURE Highest-scoring solution found
\STATE Decompose each method in $\mathcal{M}$ into ideas; initialize Idea Bank $\mathcal{B}$
\STATE \textit{(optional)} augment $\mathcal{B}$ with LLM-proposed ideas \COMMENT{Expert-Only vs.\ Augmented}
\STATE Initialize the search tree with a seed solution
\WHILE{mutation budget $b$ not exhausted}
    \STATE Select a solution node $x$ to mutate from the tree
    \IF{strategy is Bandit}
        \STATE Select idea $i \in \mathcal{B}$ using the bandit score $S_i$ \COMMENT{Eq.~\ref{eq:ucb}}
    \ELSE
        \STATE Sample idea $i$ uniformly from $\mathcal{B}$
    \ENDIF
    \STATE Insert idea $i$ into the mutation prompt; the LLM mutates $x$ into a child solution $x'$
    \STATE Evaluate $x'$ to obtain score $s'$; add $x'$ to the tree
    \STATE Decompose $x'$ into ideas; add any previously unseen ideas to $\mathcal{B}$
    \FORALL{ideas $j$ contained in $x'$}
        \STATE Update $\bar{s}_j \gets$ mean score of all solutions containing $j$; refresh rank $r_j$
    \ENDFOR
\ENDWHILE
\STATE \textbf{return} highest-scoring solution in the tree
\end{algorithmic}
\end{algorithm}

\section{Experimental Setup}
\label{experiments}

We used the scRNA-seq Batch Integration problem as the testbed and conducted experiments to investigate several key design choices for Idea Search.

\subsection{scRNA-seq Batch Integration}


Single-cell RNA sequencing (scRNA-seq) revolutionizes modern biology by quantifying transcriptomic profiles at the individual cell level, producing high-dimensional and inherently sparse datasets. However, combining data from different experiments introduces ``batch effects''—systematic technical variations that can obscure true biological heterogeneity. Batch integration algorithms typically ingest expression data from distinct experimental batches to generate a representation that mitigates technical artifacts while preserving biological information. This output can be a corrected feature matrix, a low-dimensional embedding, or a joint neighborhood graph.

For the dataset, we used CZ CELLxGENE Discover~\citep{czi2025cz} for hill climbing and applied the same filtering and processing steps as ~\citep{aygun2025ai}. The scoring process uses the metrics and protocol defined in the OpenProblems v2.0.0 batch
integration benchmark~\citep{luecken2025defining}.

We used the expert method summaries provided by~\citep{aygun2025ai}, which were extracted directly from the manuscripts of individual methods: BBKNN~\citep{polanski2020bbknn}, ComBat~\citep{johnson2007adjusting}, Harmony~\citep{korsunsky2019fast}, LIGER~\citep{liu2020jointly}, SCALEX~\citep{xiong2022online}, Scanorama~\citep{hie2019efficient}, TabVI~\citep{chandrashekar2025tabvi}, batchelor fastMNN~\citep{fastmnn}, mnnpy~\citep{haghverdi2018batch}, scVI~\citep{lopez2018deep}. We then decomposed these descriptions into individual ideas, as shown in Figure~\ref{fig:idea_search_top}, and named this collection as the ``Expert-Only'' Bank.

\subsection{Sampling Techniques} 

We treat the selection of an idea as a critical step and compare two strategies: 

\textbf{Random Sampling:} Ideas are sampled uniformly at random from the bank.

\textbf{Bandit Sampling:} 
We frame the idea selection as a multi-armed bandit problem~\citep{auer2002finite}. Given the scale differences across ideas and tasks, we adopt a rank-based scoring heuristic rather than relying on raw mean estimates. To balance the exploitation of high-performing ideas with the exploration of less-tested candidates, we calculate the score $S_i$ for each idea $i \in \{1, \dots, N\}$, where $N$ is the size of the Idea Bank, using a modified Upper Confidence Bound (UCB) strategy, defined as:

\begin{equation}
\label{eq:ucb}
    S_i = \frac{1}{r_i} + \frac{\alpha}{N} \sqrt{ \frac{\ln(1 + T)}{1 + t_i} }
\end{equation}

where $r_i$ is the rank of idea $i$ based on its current score, $\alpha$ is a hyperparameter determining the level of exploration (set to $\alpha=1$ by default), and $t_i$ denotes the number of times idea $i$ has been implemented. The term $T = \sum_{j=1}^{N} t_j$ represents the total count of implementations across the entire Idea Bank. When an idea has not yet been scored, its score is initialized to a maximum value, encouraging the Tree Search to explore all ideas at least once.

\subsection{Idea Bank Augmentation} 

We compare the effect of the initial bank's size and quality: 

\textbf{Expert-Only Bank:} A small, high-quality bank containing only ideas decomposed from human-expert methods, including a total of 30 ideas.

\textbf{Augmented Bank:} The Augmented Bank was created by supplementing the Expert-Only Bank with 49 additional ideas generated through LLM brainstorming.

\subsection{Prompting Strategies for Idea Integration} 

We study the effect of the LLM's instructions for incorporating an idea, as not all ideas are compatible with a given solution: 

\textbf{Conservative:} Prioritizes the integrity of the current solution. New ideas are only used to improve the current implementation.

\textbf{Exploratory:} Prioritizes the new idea. The LLM is instructed to attempt implementation even if it requires significant modification or refactoring of the current solution.

\section{Results}
\label{results}

We evaluate our framework on the single-cell RNA-sequencing (scRNA-seq) batch integration benchmark. Our baseline is the pure Tree Search optimization system without any Idea Bank integration. For each experiment, we conducted 5 independent trials and report both the mean and maximum scores achieved across the trials to assess both consistent performance and best-case discovery. All scores are computed on the validation split following the OpenProblems v2.0.0 protocol; we therefore report relative gains over the baseline rather than comparisons against the held-out test leaderboard.

\subsection{Efficacy of Idea Search vs. Baseline: Both Sampling Strategies Reliably Break the Plateau}

\begin{figure}[ht]
    \centering
    \begin{minipage}[c]{0.72\linewidth}
        \centering
        \begin{subfigure}{0.49\linewidth}
            \centering
            \includegraphics[width=\linewidth]{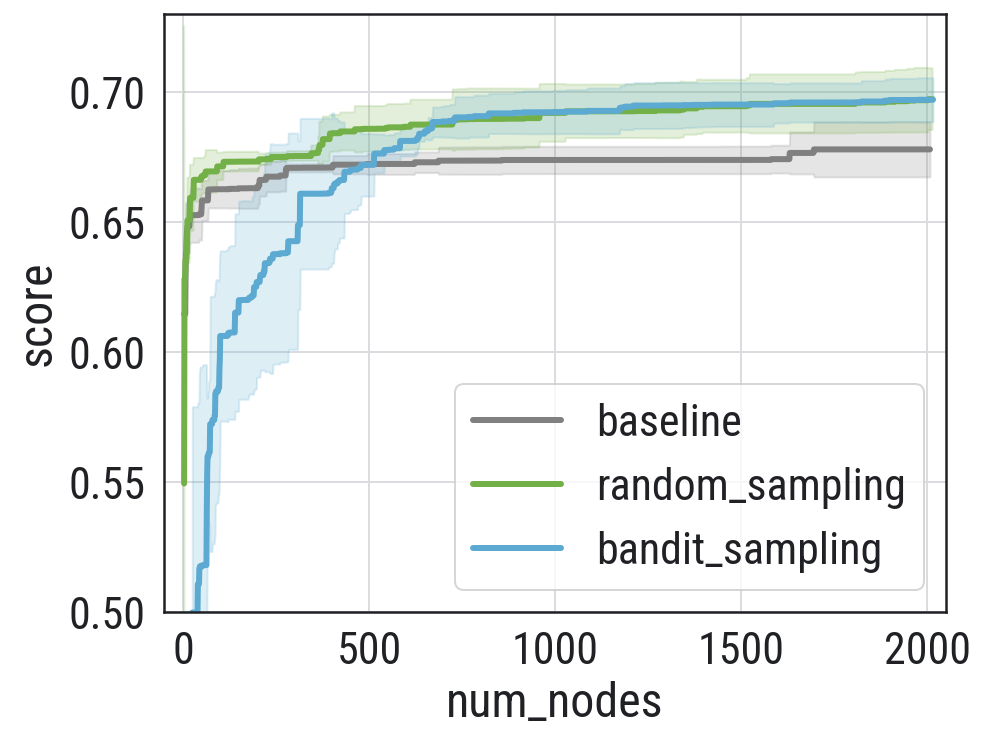}
            \caption{Mean}
            \label{fig:sampling-mean}
        \end{subfigure}
        \hfill
        \begin{subfigure}{0.49\linewidth}
            \centering
            \includegraphics[width=\linewidth]{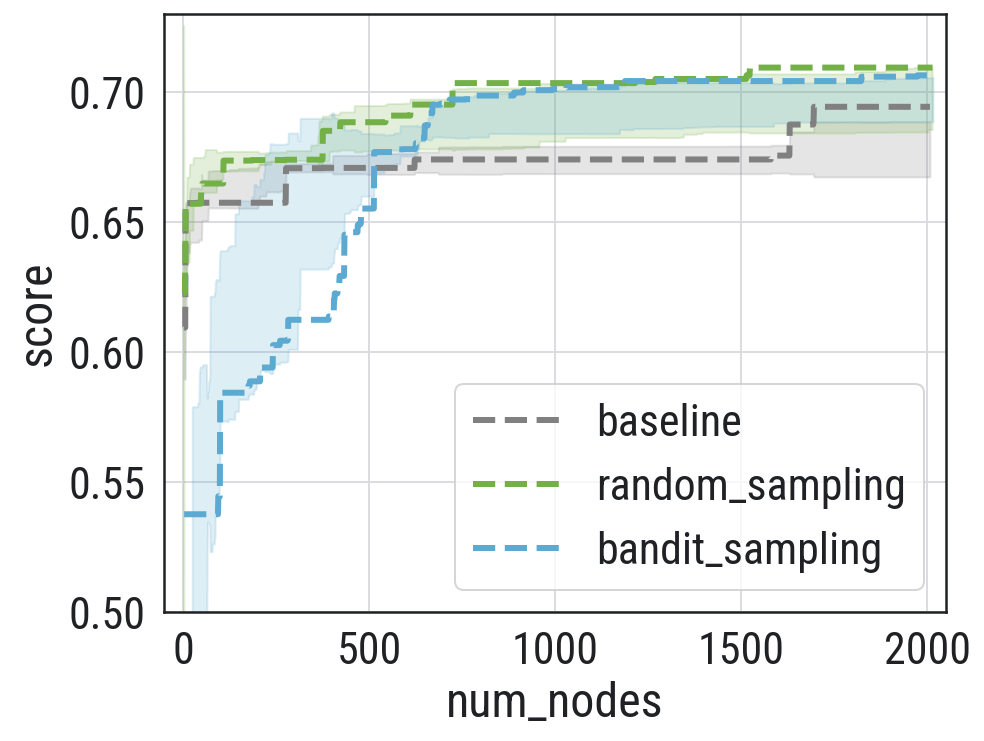}
            \caption{Max}
            \label{fig:sampling-max}
        \end{subfigure}
    \end{minipage}\hfill
    \begin{minipage}[c]{0.25\linewidth}
        \caption{\textbf{Efficacy of Idea Search vs. Baseline}: Both Random and Bandit Sampling break the baseline's plateau consistently.}
        \label{fig:sampling}
    \end{minipage}
\end{figure}

Both random and bandit idea sampling reliably break the plateau of the pure Tree Search baseline~(Fig.~\ref{fig:sampling}). The baseline, which is prompted with no ideas, typically plateaus early (around 300 nodes), reaching a mean score of 0.678 $\pm$ 0.011 and a max of 0.694. Both sampling strategies consistently outperform it, breaking the plateau relatively early (around 500 nodes). The bandit method shows a slightly slower initial hill-climb---expected, given its forced initial exploration phase where it tests all available ideas---but both strategies reach similar final scores by mean (random: 0.697 $\pm$ 0.012, bandit: 0.697 $\pm$ 0.008) and max (random: 0.709, bandit: 0.706).

\subsection{Effect of Idea Bank Augmentation: Helps Bandit Sampling but not Random}

\begin{figure}[ht]
    \centering
    \begin{minipage}[c]{0.72\linewidth}
        \centering
        \begin{subfigure}{0.95\linewidth}
            \centering
            \includegraphics[width=\linewidth]{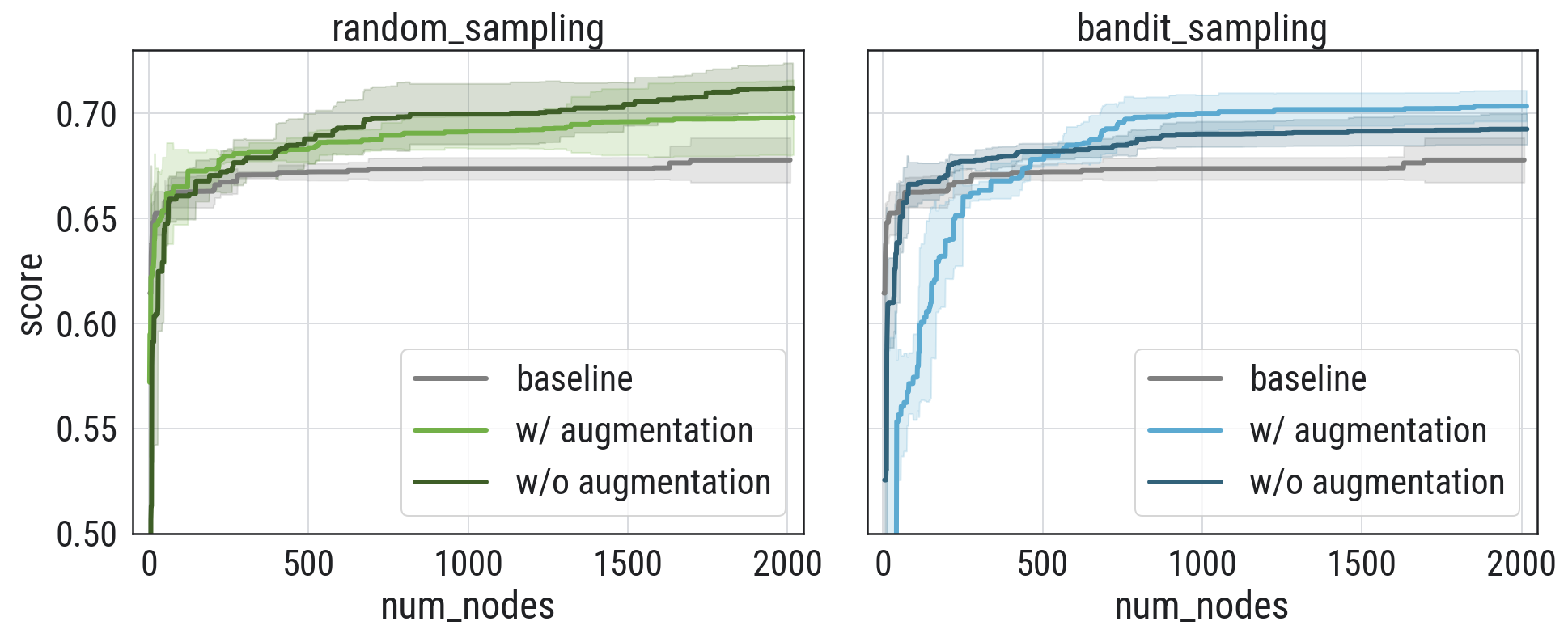}
            \caption{Mean}
            \label{fig:idea-bank-mean}
        \end{subfigure}
        \par\medskip
        \begin{subfigure}{0.95\linewidth}
            \centering
            \includegraphics[width=\linewidth]{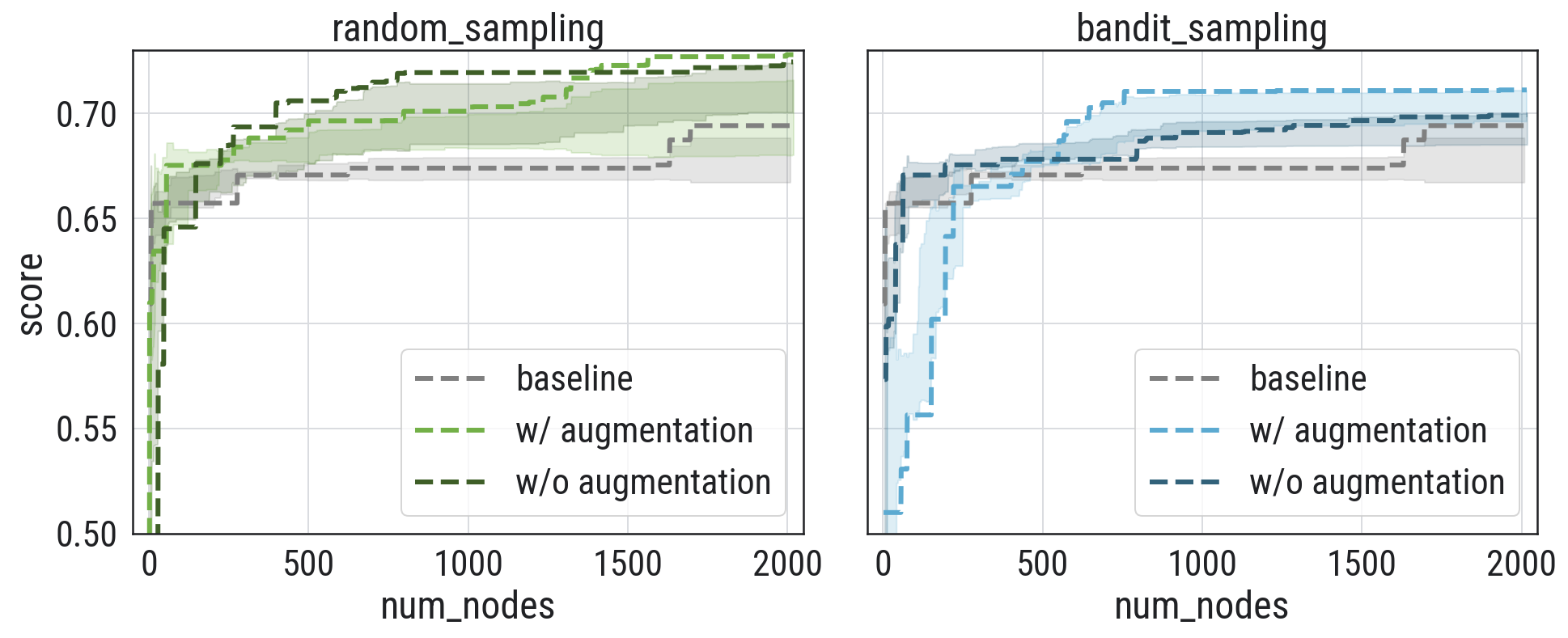}
            \caption{Max}
            \label{fig:idea-bank-max}
        \end{subfigure}
    \end{minipage}\hfill
    \begin{minipage}[c]{0.25\linewidth}
        \footnotesize
        \caption{\textbf{Effect of Idea Bank Augmentation:} Bandit Sampling benefits from augmentation, while Random Sampling gets higher score on average with the original ``Expert-Only'' Idea Bank.}
        \label{fig:augmentation}
    \end{minipage}
\end{figure}

We next analyzed the effect of Idea Bank augmentation, revealing a mixed effect that was highly dependent on its interaction with the sampling strategy~(Fig.~\ref{fig:augmentation}). With random sampling, the smaller, expert-only Idea Bank consistently performed better on average (0.712 $\pm$ 0.012). The augmented bank, while producing lower mean scores (0.698 $\pm$ 0.018), did occasionally produce a single, very high-scoring outlier solution with score 0.728. The bandit sampling strategy consistently benefited from the augmented bank, with the mean score improved from 0.692 $\pm$ 0.007 to 0.703 $\pm$ 0.007. 
This suggests that the bandit sampling's score-guided mechanism is particularly well-suited for the augmented Idea Bank. Its ability to intelligently navigate the larger conceptual space creates a synergistic effect, where the larger bank becomes an advantage. In contrast, pure random sampling benefits from a more constrained, high-quality space to avoid sampling low-quality ideas too frequently.

\subsection{Effect of Prompting and Exploration: Exploratory Prompting Surfaces the Best Solutions, while More Sampling-Level Exploration Hurts}

\begin{figure}[ht]
    \centering
    \begin{minipage}[c]{0.72\linewidth}
        \centering
        \begin{subfigure}{0.95\linewidth}
            \centering
            \includegraphics[width=\linewidth]{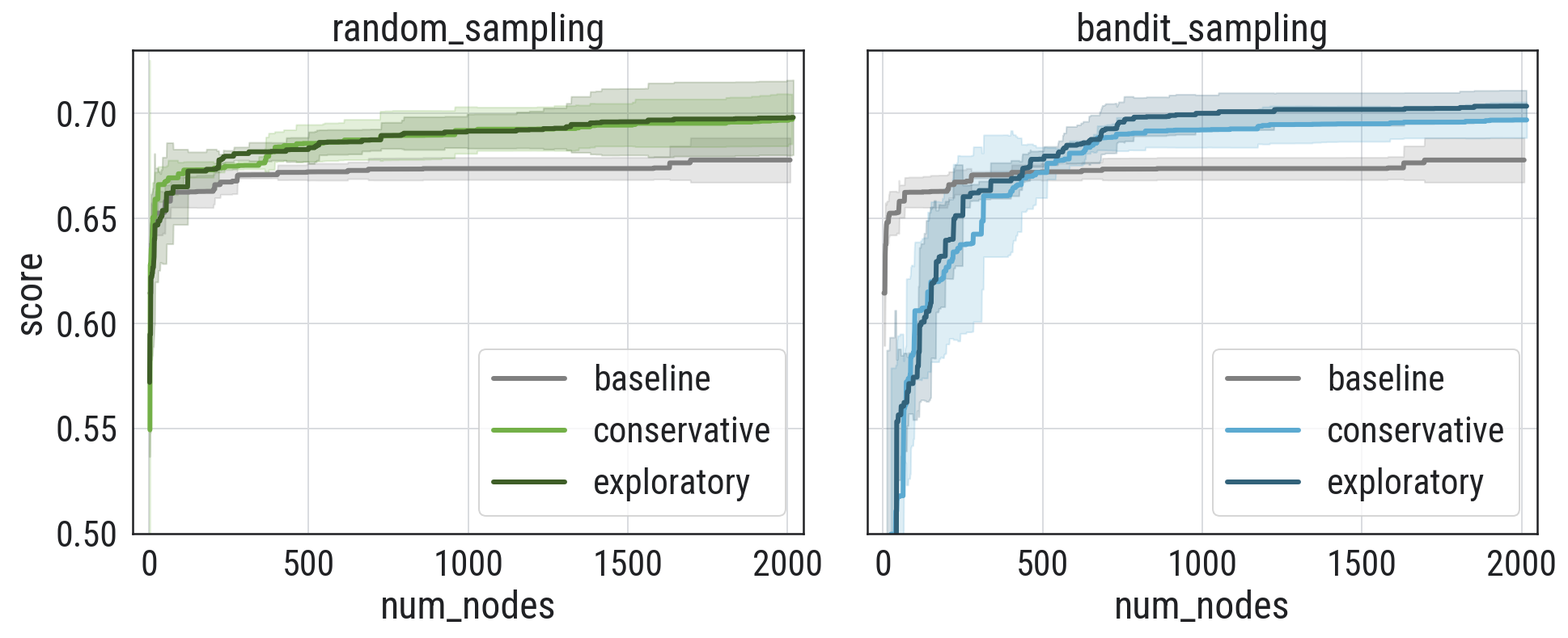}
            \caption{Mean}
            \label{fig:prompting-mean}
        \end{subfigure}
        \par\medskip
        \begin{subfigure}{0.95\linewidth}
            \centering
            \includegraphics[width=\linewidth]{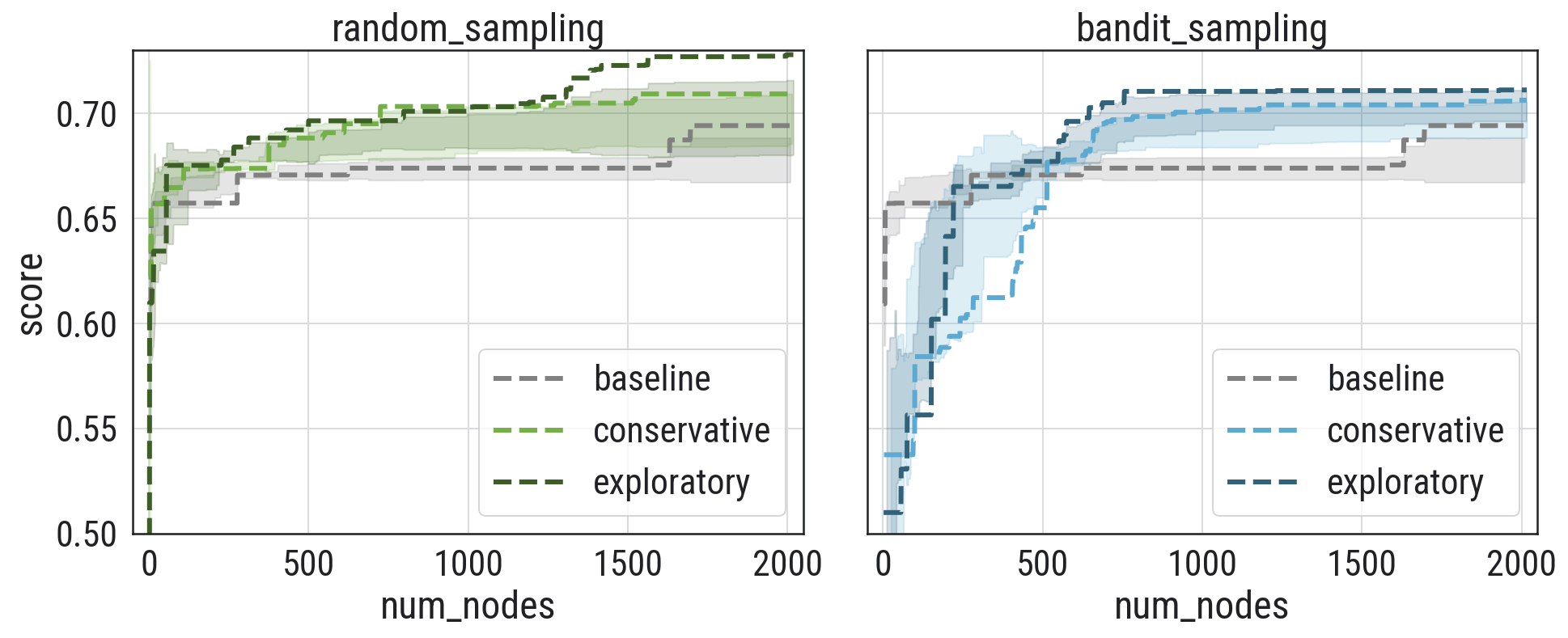}
            \caption{Max}
            \label{fig:prompting-max}
        \end{subfigure}
    \end{minipage}\hfill
    \begin{minipage}[c]{0.25\linewidth}
        \footnotesize
        \caption{\textbf{Effect of Prompting Strategy:} Exploratory Prompting is generally beneficial, and when combined with Random Sampling, it leads to the discovery of an exceptional outlier solution.}
        \label{fig:prompting}
    \end{minipage}
\end{figure}

Exploratory prompting helps mainly by enabling rare, exceptional solutions rather than by raising the mean~(Fig.~\ref{fig:prompting}). Comparing conservative (prioritize the current implementation) versus exploratory (prioritize new ideas) prompting, mean scores were comparable across strategies (around 0.697), with the exploratory prompt showing only a marginal mean advantage under bandit sampling (0.703 $\pm$ 0.007). The maximum scores tell the real story: the ``exploratory'' prompt paired with random sampling yielded the single highest-scoring solution with score 0.728. This suggests that exploratory prompting--one that attempts new ideas more aggressively--when combined with random sampling, creates a ``high-risk, high-reward'' scenario, encouraging the discovery of exceptional outlier solutions.

\begin{figure}[ht]
    \centering
    \begin{minipage}[c]{0.72\linewidth}
        \centering
        \begin{subfigure}{0.49\linewidth}
            \centering
            \includegraphics[width=\linewidth]{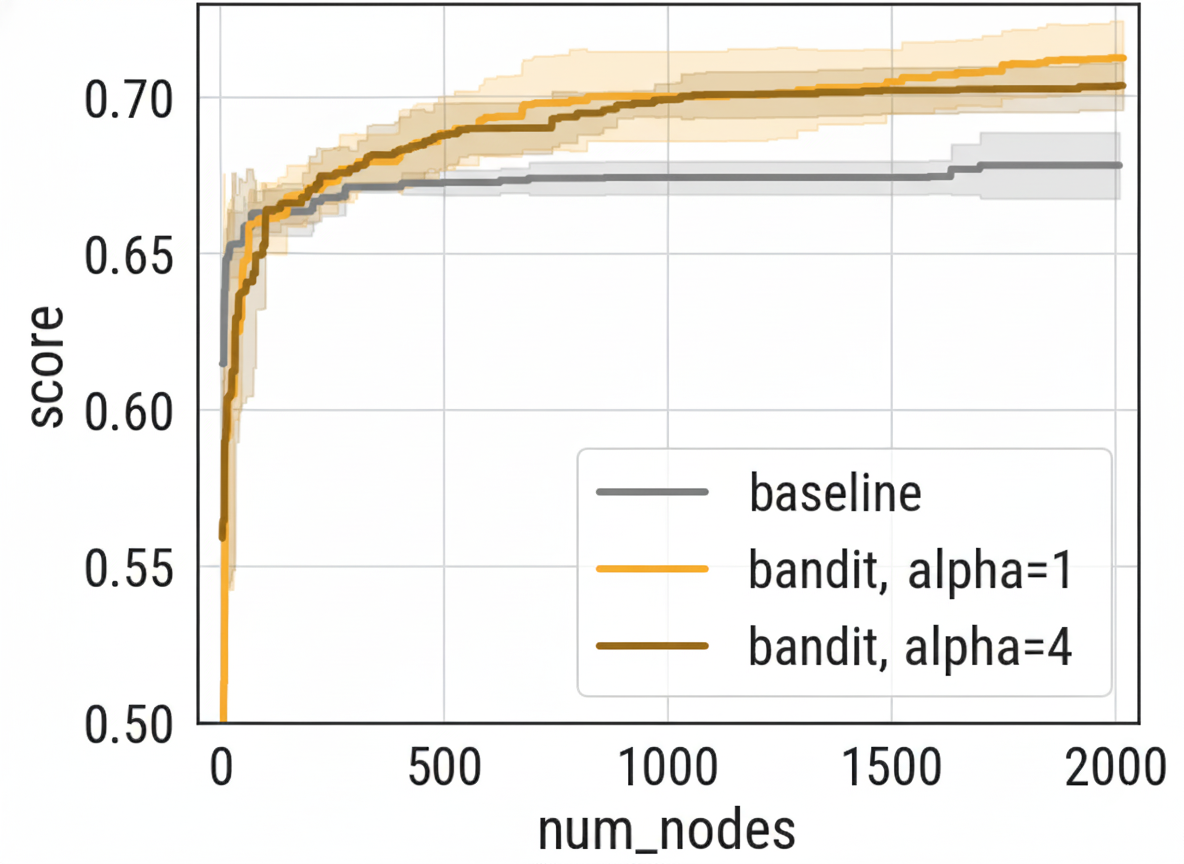}
            \caption{Mean}
            \label{fig:alpha-mean}
        \end{subfigure}
        \hfill
        \begin{subfigure}{0.49\linewidth}
            \centering
            \includegraphics[width=\linewidth]{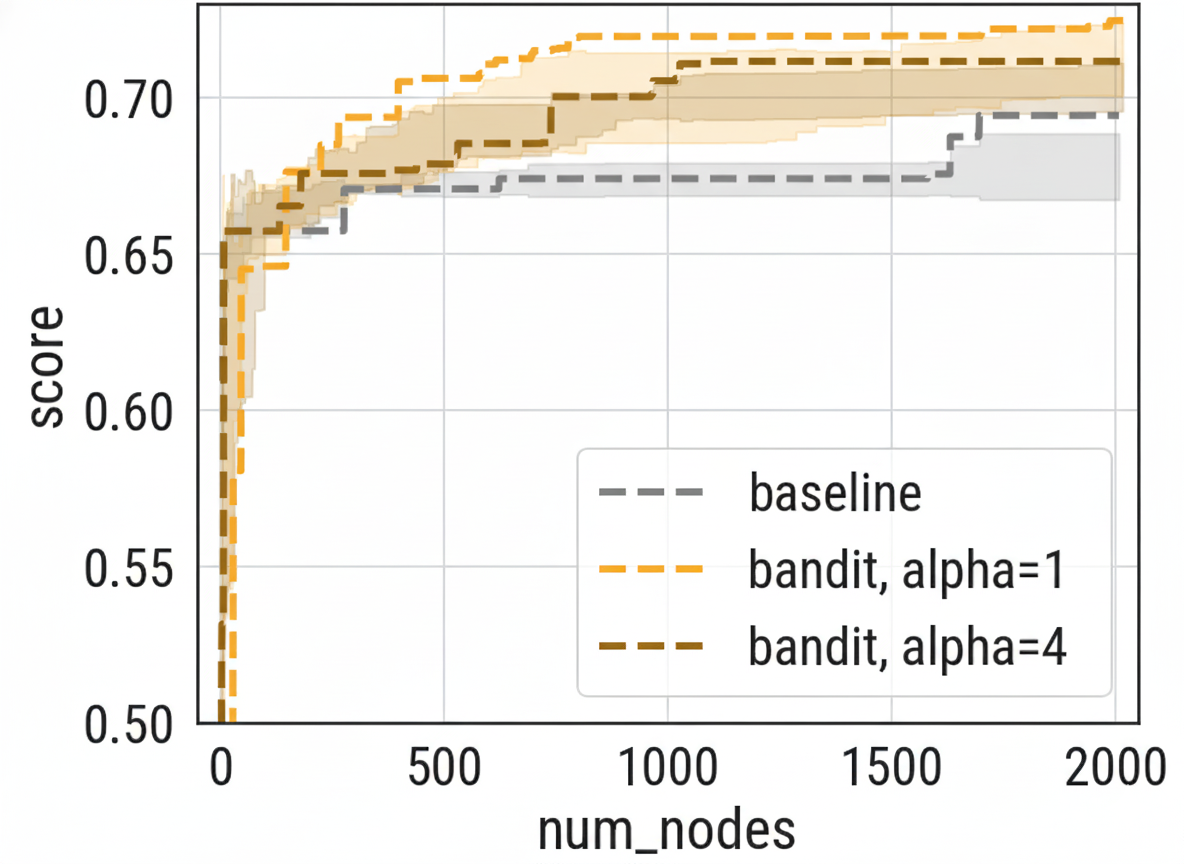}
            \caption{Max}
            \label{fig:alpha-max}
        \end{subfigure}
    \end{minipage}\hfill
    \begin{minipage}[c]{0.25\linewidth}
        \caption{\textbf{Effect of the Exploration Parameter in Bandit Sampling:} An increased exploration parameter in Bandit Sampling surprisingly decreases performance.}
        \label{fig:alpha}
    \end{minipage}
\end{figure}

Given that implementation-level exploration was beneficial, we tested if sampling-level exploration was also universally beneficial. In the bandit sampling algorithm, the alpha value controls the exploration-exploitation trade-off. We conducted experiments comparing the default (alpha=1) with a high-exploration value (alpha=4). Surprisingly, we observed that a high alpha value was detrimental to the optimization process, leading to a decrease of the mean score from 0.712 $\pm$ 0.012 to 0.703 $\pm$ 0.008~(Fig.~\ref{fig:alpha}).

These combined outcomes suggest that the mechanism of exploration is critical. While implementation-level exploration (via prompting) was effective at discovering top solutions, a naive increase in sampling-level exploration (high alpha) was counterproductive.

\textbf{Scope of the results.} Across all configurations, injecting ideas produces a consistent but small upward shift in the mean (baseline 0.678 to roughly 0.70) alongside a heavier upper tail (best 0.728 versus the baseline's 0.694). The gains are modest relative to trial-to-trial variance: the standard deviation across 5 trials (0.008--0.018) is comparable to the mean improvement over the baseline (about 0.02), so the practical signal sits in the best-of-$n$ solution more than in the average run. The design choices change \emph{where} that signal appears---augmentation pairs with bandit sampling, exploratory prompting under random sampling produces the single best solution, and higher sampling-level exploration ($\alpha=4$) hurts---rather than uniformly raising the mean. All effects are measured on one validation-split benchmark with a single backbone LLM (Gemini 2.5 Pro); we read them as evidence that injecting ideas breaks the plateau and reshapes the score distribution, not as a calibrated estimate of effect size.

\section{Discussion}
\label{conclusion}

This work introduced Idea Search, a novel framework that systematically integrates conceptual exploration into the Tree Search optimization process for automated scientific discovery. By developing a dynamic Idea Bank that guides code mutation, we directly addressed the core limitation of pure Tree Search: its propensity to get trapped in local optima. Applied to scRNA-seq batch integration, Idea Search broke the performance plateau established by a strong Tree Search baseline (mean 0.678 to 0.697), supporting the value of systematically injecting execution-verified ideas---though, as our analysis shows, the size of the gain depends on how the Idea Bank is sampled, augmented, and prompted.

\textbf{Limitations.} Our evidence is narrow in scope. All results come from a single task with one backbone LLM (Gemini 2.5 Pro), so whether Idea Search transfers to other domains or models is untested. Scores are computed on the OpenProblems validation split; we report gains relative to our own baseline and make no claim against the held-out test leaderboard. With five trials per configuration and absolute gains (about 0.02) on the order of the trial-to-trial spread, our comparisons identify the direction of each effect but not a precise effect size. Finally, the Idea Bank is built through LLM summarization: an idea's score is the average over every solution that used it, which assigns credit coarsely when several ideas co-occur in one solution and inherits any bias in how the LLM decomposes methods into ideas.

Future work can focus on three directions. First, research ideation systems such as Deep Research~\citep{deepresearch_g, deepresearch_o} and others~\citep{baek2024researchagent, wang2024scimon, li2024chain, hu2024nova} could supply external ideas, enriching the bank. Second, extending Idea Search to other scientific domains would test its generality. Finally, since the gains concentrate in rare outliers rather than the mean, future work should investigate solution variance, reducing trial-to-trial noise while preserving the capacity for rare, high-impact discoveries.

\clearpage



\bibliography{example_paper}
\bibliographystyle{colm2026_conference}

\newpage
\appendix
\section{Appendix}

\subsection{Prompts for Generating and Augmenting the Idea Bank}

\begin{tcolorbox}[
    colframe=blue!75!black,
    colback=blue!5!white,
    boxrule=0.5pt, 
    title=\textbf{Generate the Idea Bank}
]
For a single-cell batch integration problem, there is a list of existing methods. And I would like you to analyze them, decompose those methods into a list of unitary, factored ideas so that I can sample and compose them to invent new methods. Each idea should be a succinct phrase and the list should be comprehensive. Here are the existing methods:

\{ List of Expert Methods \}

\end{tcolorbox}

\begin{tcolorbox}[
    colframe=blue!75!black,
    colback=blue!5!white,
    boxrule=0.5pt, 
    title=\textbf{Augment the Idea Bank}
]
For a single-cell batch integration problem, there is a list of ideas that can be composed and applied to this problem. And I would like you to brainstorm and suggest more relevant ideas. Here are the existing methods:

\{ Original Expert-Only Idea Bank \}

\end{tcolorbox}

\subsection{Idea Bank}

\begin{tcolorbox}[
    colframe=blue!75!black,
    colback=blue!5!white,
    boxrule=0.5pt, 
    title=\textbf{Expert-Only Bank}
]
Use an iterative refinement strategy

Regularize latent space with KL divergence

Merge per batch nearest neighbor sets

Use a batch agnostic encoder and batch aware decoder

Match cells by shared factor loading similarity

Decompose latent space by variation source

Integrate in a linear embedding like PCA or SVD

Merge batches sequentially to a reference

Decompose into shared and dataset specific factors

Generate an integrated neighborhood graph

Use joint matrix factorization

Correct in high dimensional gene space

Condition decoder on batch ID

Integrate by modifying a KNN graph

Apply iterative cluster centroid-based correction

Integrate on metagene loadings

Use empirical Bayes to stabilize gene corrections

Learn per sample gene weights via attention

Use Mutual Nearest Neighbors as anchors

Model additive and multiplicative batch effects

Use diversity penalized soft clustering

Compute corrections between all dataset pairs

Use a Zero Inflated Negative Binomial loss

Preprocess using log transform or cosine normalization

Employ a Variational Autoencoder framework

Quantile normalize factor loadings across batches

Align implicitly via mixed batch training

Apply correction from weighted MNN vectors

Integrate via a nonlinear latent space

Use per batch normalization layers like DSBN
\end{tcolorbox}

\begin{tcolorbox}[
    colframe=blue!75!black,
    colback=blue!5!white,
    boxrule=0.5pt, 
    breakable,
    title=\textbf{Augmented Bank}
]

Transformer encoder architecture applied to feature vectors \\
Attention mechanisms within an autoencoder \\
Generate an integrated neighborhood graph \\
Merge batches sequentially to a reference \\
Maximum Mean Discrepancy loss on the latent space \\
Graph Neural Networks on the joint graph \\
Triplet loss to pull similar points together \\
Normalizing flows to learn an invertible transformation \\
Decompose into shared and dataset-specific factors \\
Cycle-Consistent GANs for data style transfer \\
Generative diffusion models conditioned on source \\
Correct in high-dimensional gene space \\
Adversarial Gradient Reversal Layer in a neural network \\
Use diversity-penalized soft clustering \\
Contrastive learning for a batch-invariant space \\
Bootstrap aggregating of correction models for robustness \\
Align implicitly via mixed-batch training \\
Adversarial discriminator on the latent representation \\
Conformal prediction for uncertainty-aware integration \\
Bayesian structure learning to model data generation \\
Multi-scale analysis using wavelet transforms \\
Building a joint k-Nearest Neighbor graph \\
Minimizing mutual information between latent space and source ID \\
Kernel Principal Component Analysis on a joint similarity matrix \\
Center loss for increasing intra-class compactness in latent space \\
Anchor-based correction using a defined set of control points \\
Topological Data Analysis to align manifold structures \\
Data augmentation by interpolating between samples in latent space \\
Model additive and multiplicative batch effects \\
Self-supervised learning with a pretext task \\
Meta-learning a general-purpose correction model \\
Diffusion map embedding on a joint affinity matrix \\
Use a batch-agnostic encoder and batch-aware decoder \\
Non-negative Matrix Factorization \\
Disentangled representation learning \\
Joint matrix factorization \\
Log-transform the data \\
Alternating Direction Method of Multipliers for optimization \\
Deep Correlation Maximization \\
Embedding into Hyperbolic space for hierarchical data \\
Semi-supervised domain adaptation using partial labels \\
Feature selection for highly variable features \\
Use Sliced-Wasserstein distance as loss \\
Causal inference modeling to estimate and remove confounder effects \\
Mutual Nearest Neighbor pairing for alignment \\
Quantile-normalize factor loadings across batches \\
UMAP or t-SNE embedding on the joint graph \\
Latent space optimization without an explicit encoder \\
Autoencoder with a shared bottleneck layer \\
Siamese networks to learn a source-invariant metric \\
Modeling continuous technical covariates with a GLM \\
Empirical Bayes harmonization
Rank-based inverse normal transformation \\
Integrate on metagene/factor loadings \\
Regressing out the source ID from principal components \\
Gromov-Wasserstein Optimal Transport for alignment \\
Quantile normalization across all datasets \\
Z-score features within each source dataset \\
Conditional VAE using the source ID as a condition \\
Wasserstein Autoencoder with an MMD penalty \\
Decompose latent space by variation source \\
Iterative centroid alignment between clusters \\
Linear mixed models with source as a random effect \\
Use per-batch normalization layers like DSBN \\
Canonical Correlation Analysis \\
Perform global one-shot optimization \\
Minimizing mutual information between latent space and source ID \\
Condition decoder on batch ID
Partial Least Squares  \\
Discriminant Analysis
Match cells by shared factor loading similarity \\
Principal Component Analysis \\
Federated learning to train a shared model without moving data \\
Regressing out the source ID from principal components \\
Use a Zero-Inflated Negative Binomial loss

Ensemble modeling by averaging multiple integration results

Regularize latent space with KL divergence

Compute corrections between all dataset pairs

Sparse Principal Component Analysis for interpretable factors

\end{tcolorbox}

\subsection{Prompts for Idea Integration}

Conservative

\begin{tcolorbox}[
    colframe=blue!75!black,
    colback=blue!5!white,
    boxrule=0.5pt, 
    title=\textbf{Conservative prompt relying on the LLM's implicit judgement for which ideas to focus on.}
]
Consult the Idea Bank below for potential improvements. Try to use the ideas to improve your current plan. If the Idea Bank is empty, you can also devise your own expert strategy. Try to apply the ideas as much as possible. 

\{ Sampled Ideas from Idea Bank \}
\end{tcolorbox}

Exploratory
\begin{tcolorbox}[
    colframe=blue!75!black,
    colback=blue!5!white,
    boxrule=0.5pt, 
    title=\textbf{Exploratory prompt to encourage LLM to focus on new ideas.}
]
Use the Idea Bank below to create a new plan. Your goal is to incorporate these ideas as much as possible, along with any lessons learned from previous revisions. No need to stick to the current code; prioritize fully implementing the new strategies from the Idea Bank. If an idea is applicable, commit to using it faithfully to achieve its full potential, no matter how complex it seems.

\{ Sampled Ideas from Idea Bank \}
\end{tcolorbox}



\end{document}